\documentclass[letterpaper, 10 pt, conference]{ieeeconf}
\IEEEoverridecommandlockouts                              
\usepackage{times}
\usepackage{epsfig}
\usepackage{graphicx}
\usepackage{amsmath}
\usepackage{amssymb}

\usepackage{graphicx}
\usepackage{epsfig}
\usepackage{epic,eepic}
\usepackage{times}
\usepackage{mathptmx}
\usepackage{amsfonts}
\usepackage{amsmath}
\usepackage{cite}
\usepackage{color}

\usepackage{times}

\definecolor{red}{rgb}{1,0,0}
\definecolor{green}{rgb}{0,1,0}
\definecolor{blue}{rgb}{0,0,1}
\definecolor{violet}{rgb}{1,0,1}
\definecolor{cyan}{cmyk}{1,0,0,0}
\definecolor{magenta}{cmyk}{0,1,0,0}
\definecolor{yellow}{cmyk}{0,0,1,0}

\definecolor{white}{rgb}{1,1,1}

\usepackage{jumoline}

\newcommand{\CO}[1]{}

\newcommand{\CommentOut}[1]{}

\newcommand{\noeditage}[1]{#1} \newcommand{\editage}[1]{}

\newcommand{\FIG}[3]{
\begin{minipage}[b]{#1cm}
\begin{center}
\includegraphics[width=#1cm]{#2}\\
{\scriptsize #3}
\end{center}
\end{minipage}
}

\newcommand{\FIGR}[3]{
\begin{minipage}[b]{#1cm}
\begin{center}
\includegraphics[angle=-90,clip,width=#1cm]{#2}
\\
{\scriptsize #3}
\vspace*{1mm}
\end{center}
\end{minipage}
}

\newcommand{\acprPaperID}{25}

\begin{document}

\author{\\
\\
\\
{\tt\small ~}
\and
\\
\\
\\
{\tt\small ~}
}

\newcommand{\SW}[2]{#1}

\newcommand{\figA}{
\begin{figure}[t]
  \begin{center}
\FIGR{7}{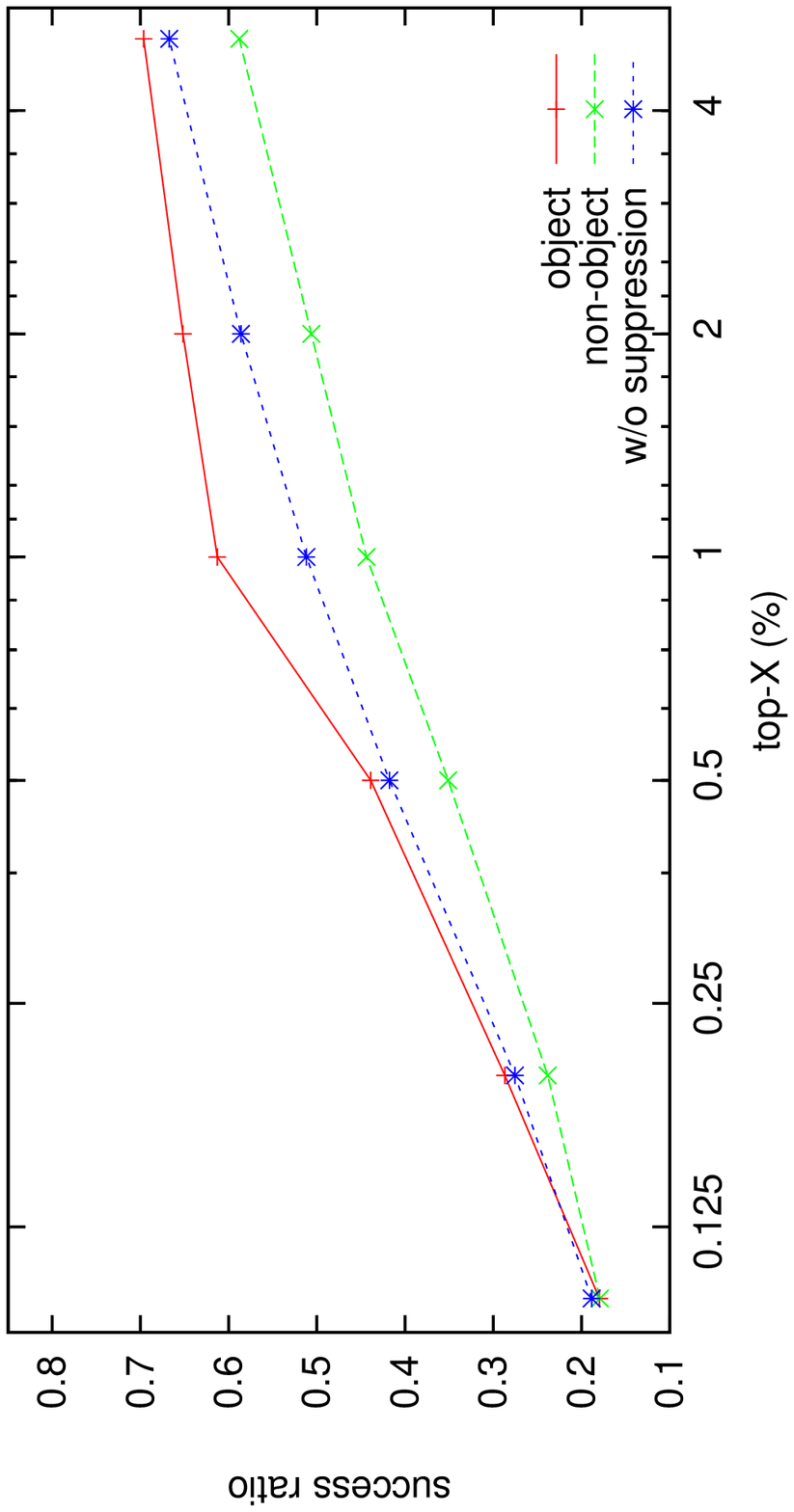}{}\vspace*{-5mm}\\
\caption{Change detection performance.}\label{fig:A}
\vspace*{-5mm}
\end{center}
\end{figure}
}

\newcommand{\figAa}{
\begin{figure}[t]
  \begin{center}
\FIGR{7}{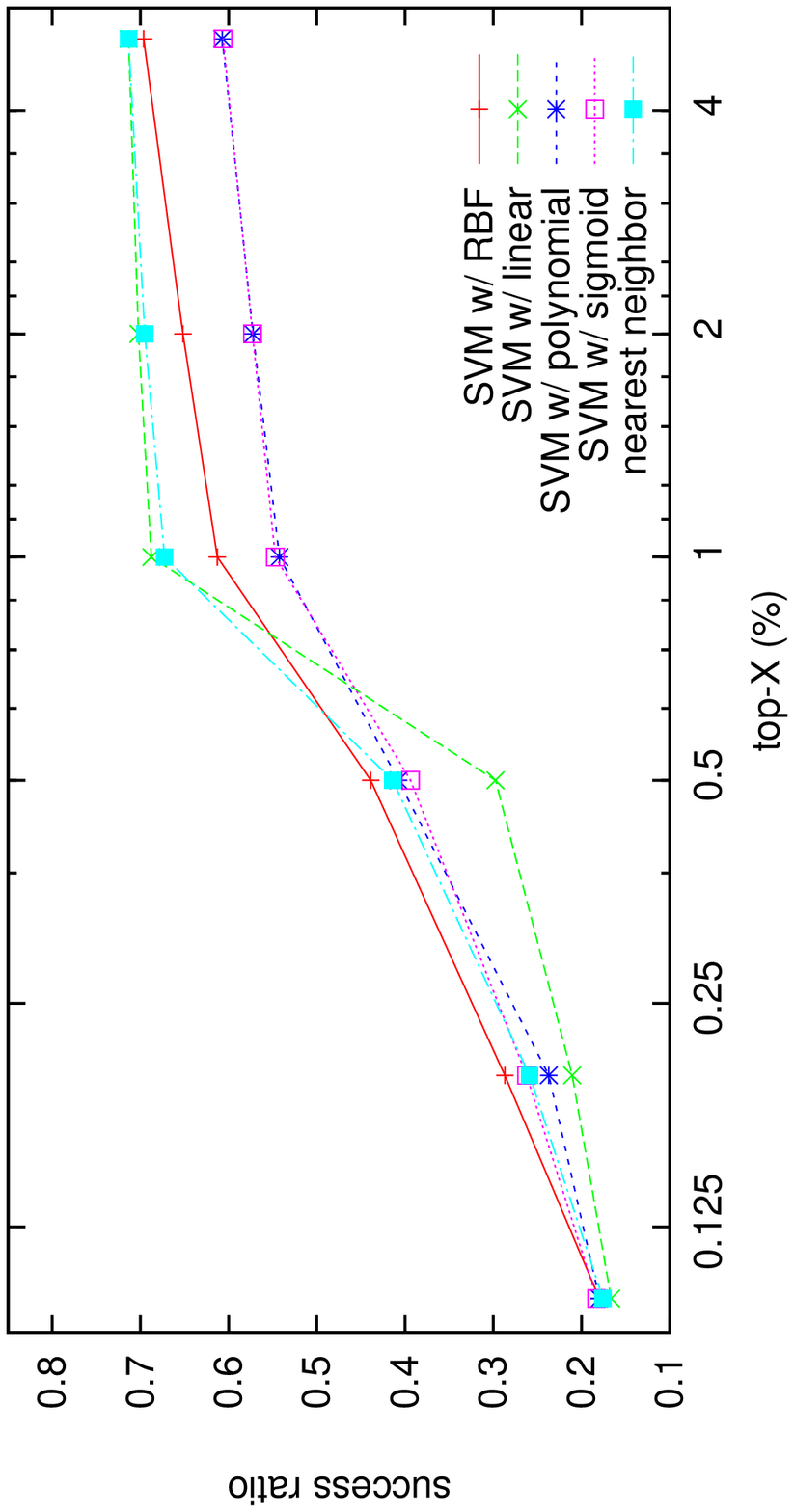}{}\vspace*{-5mm}\\
\caption{Influence of classifiers and kernels.}\label{fig:Aa}
\vspace*{-5mm}
\end{center}
\end{figure}
}

\newcommand{\figAb}{
\begin{figure}[t]
  \begin{center}
\FIGR{7}{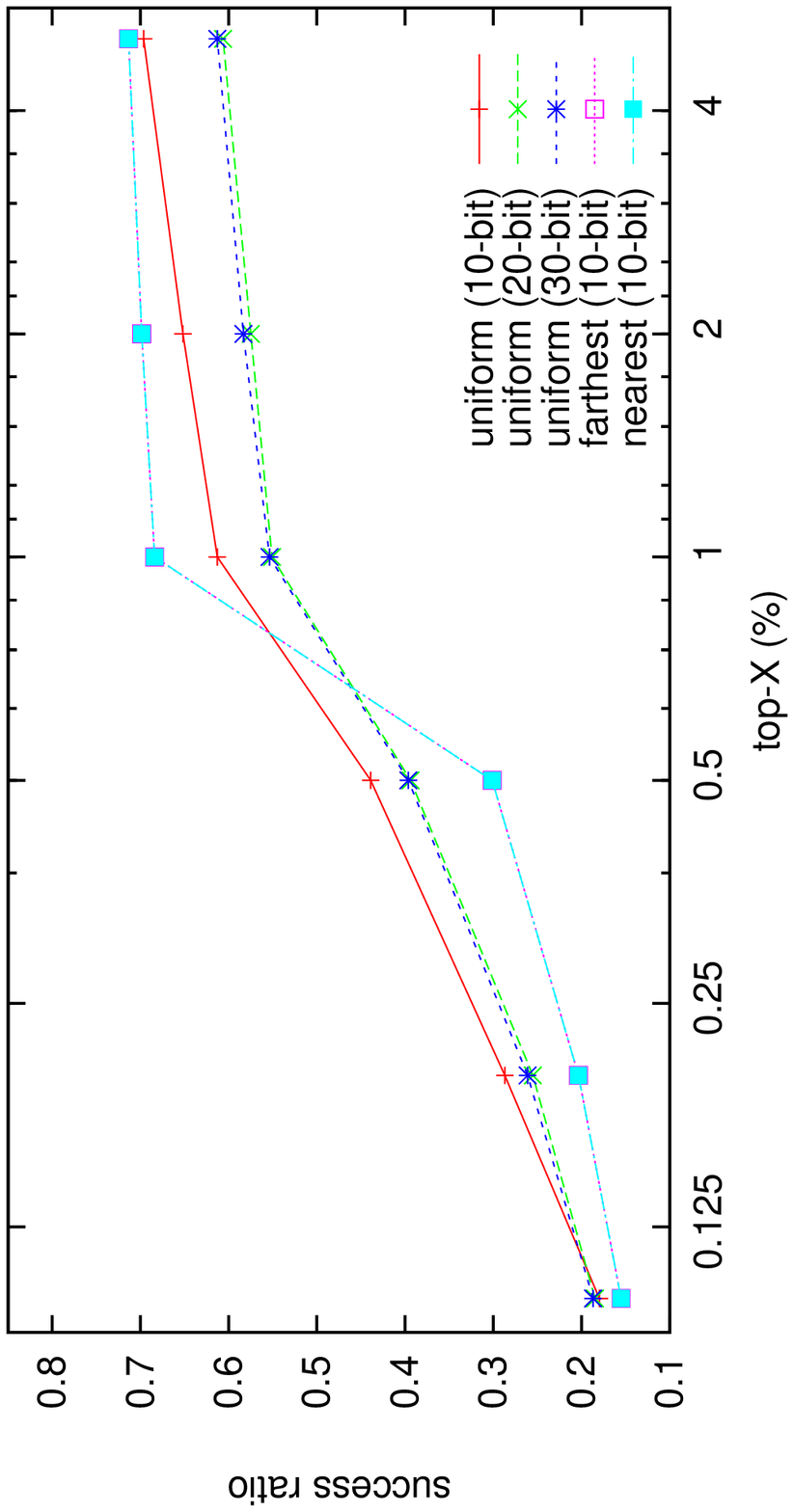}{}\vspace*{-5mm}\\
\caption{Influence of change-mining strategies.}\label{fig:Ab}
\vspace*{-5mm}
\end{center}
\end{figure}
}

\newcommand{\figAc}{
\begin{figure}[t]
  \begin{center}
\FIGR{7}{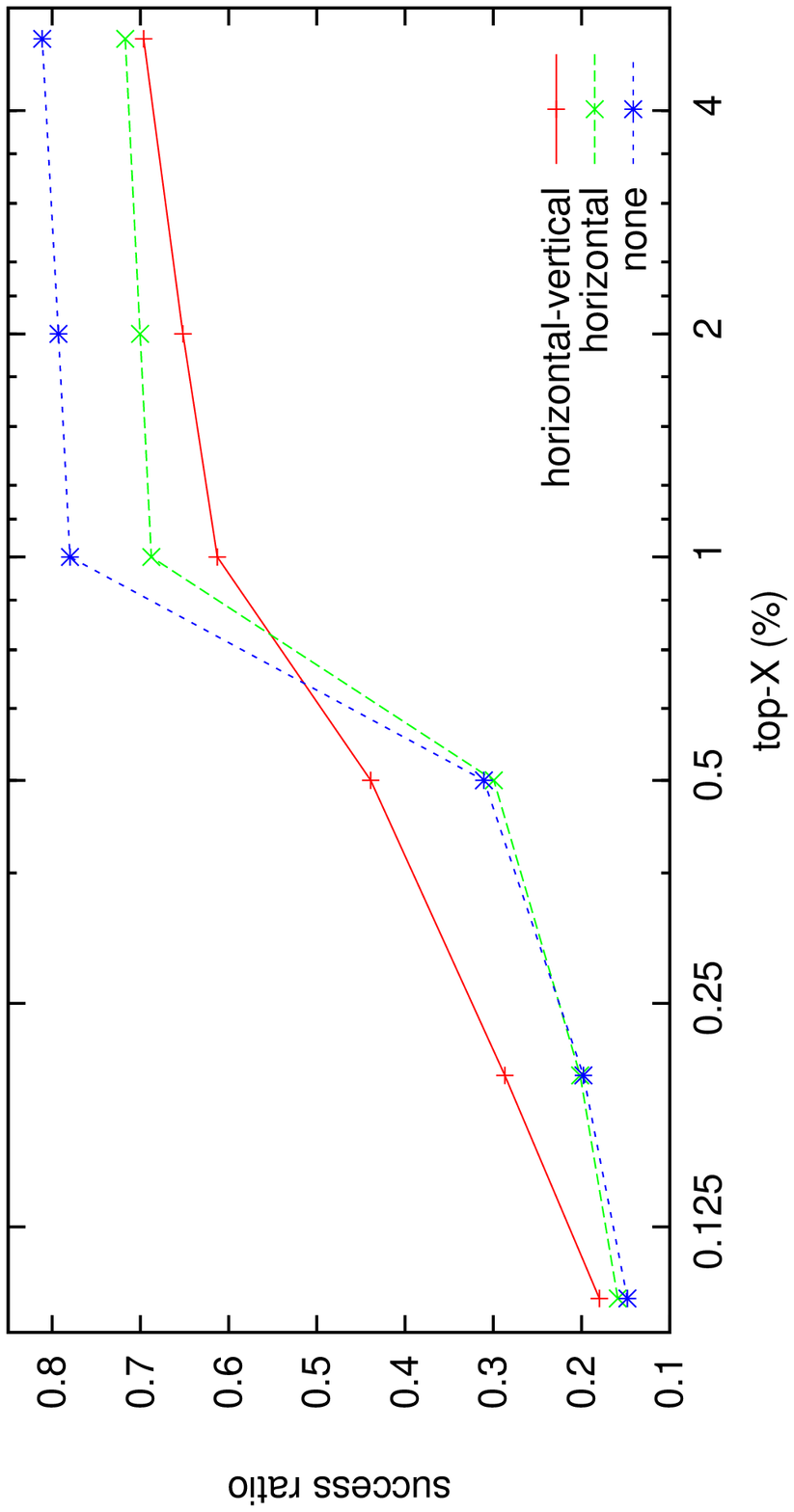}{}\vspace*{-5mm}\\
\caption{Influence of visibility-analysis strategies.}\label{fig:Ac}
\vspace*{-5mm}
\end{center}
\end{figure}
}

\newcommand{\figAd}{
\begin{figure}[t]
  \begin{center}
\FIGR{7}{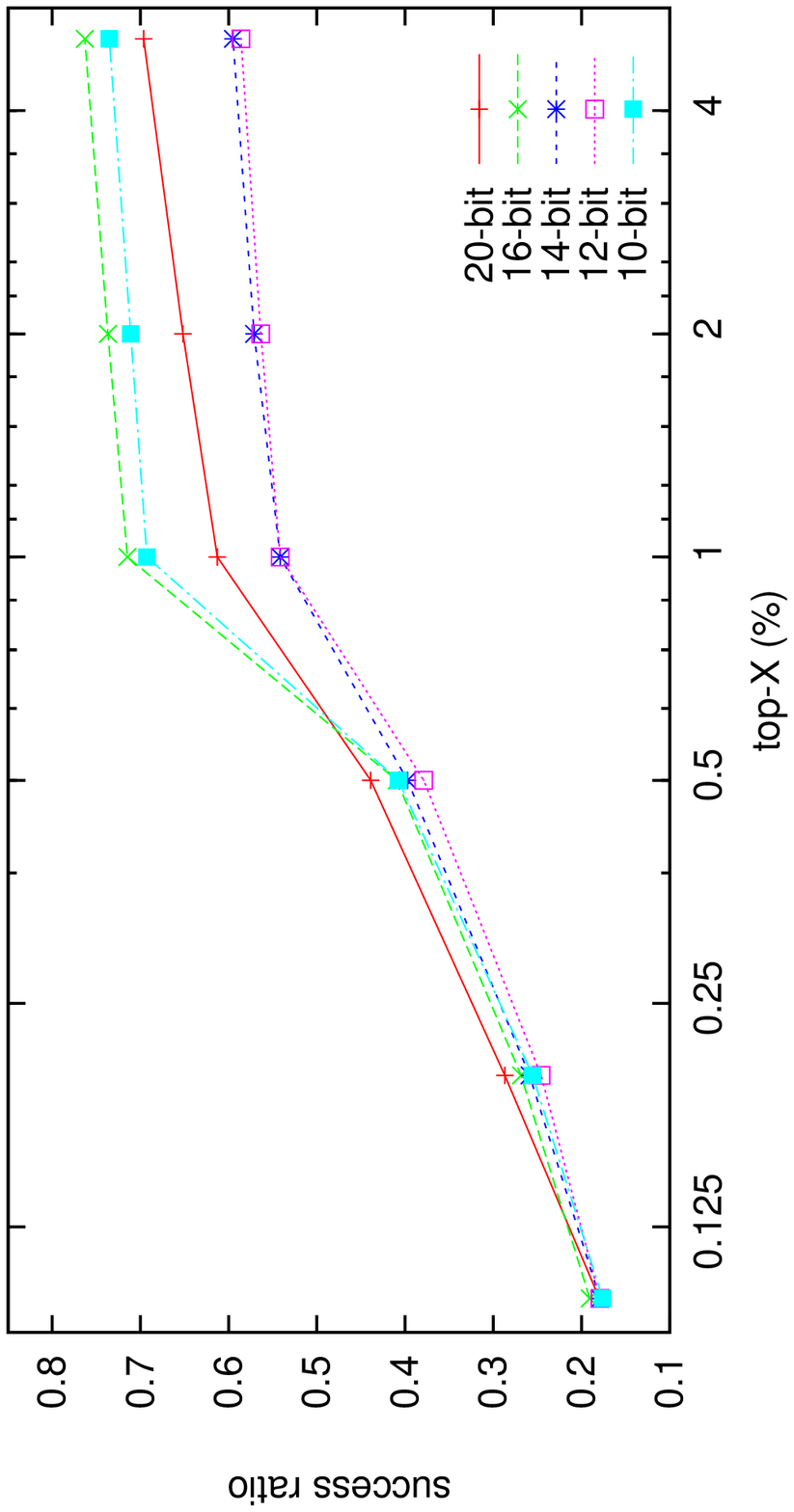}{}\vspace*{-5mm}\\
\caption{Influence of vocabulary size.}\label{fig:Ad}
\vspace*{-5mm}
\end{center}
\end{figure}
}

\newcommand{\figB}{
\begin{figure}[t]
  \begin{center}
\FIG{8}{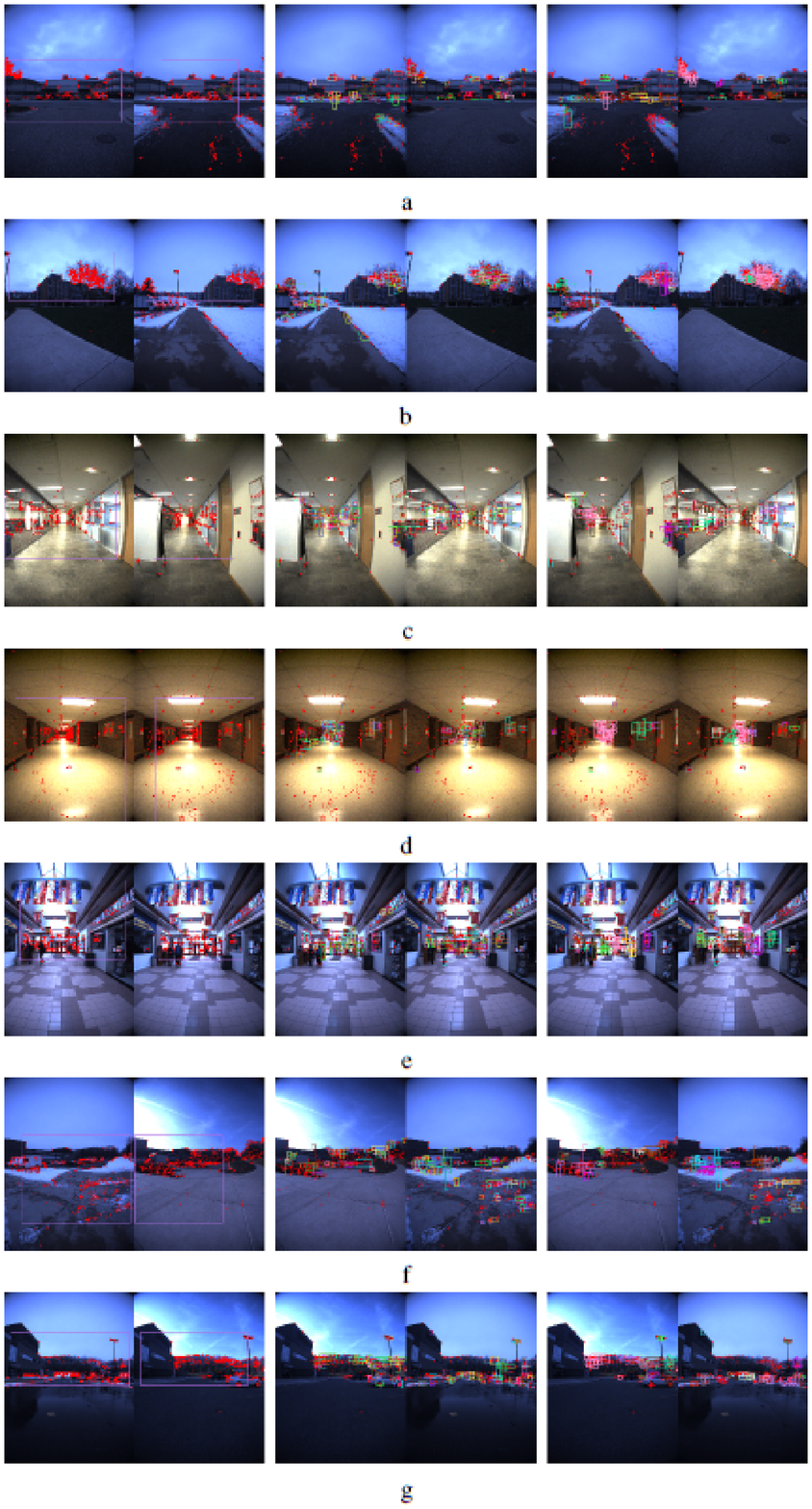}{}
\caption{Examples of scene recognition.
Left panel:
ORB keypoints (red dots)
and commonly visible regions
estimated between query and reference images (purple boxes).
Middle and right panels:
Object regions and object clusters extracted from both images,
using different colors for different regions and clusters.
}\label{fig:B}
\vspace*{-5mm}
  \end{center}
\end{figure}
}

\newcommand{\figC}{
\begin{figure}[t]
  \begin{center}
\FIG{7.5}{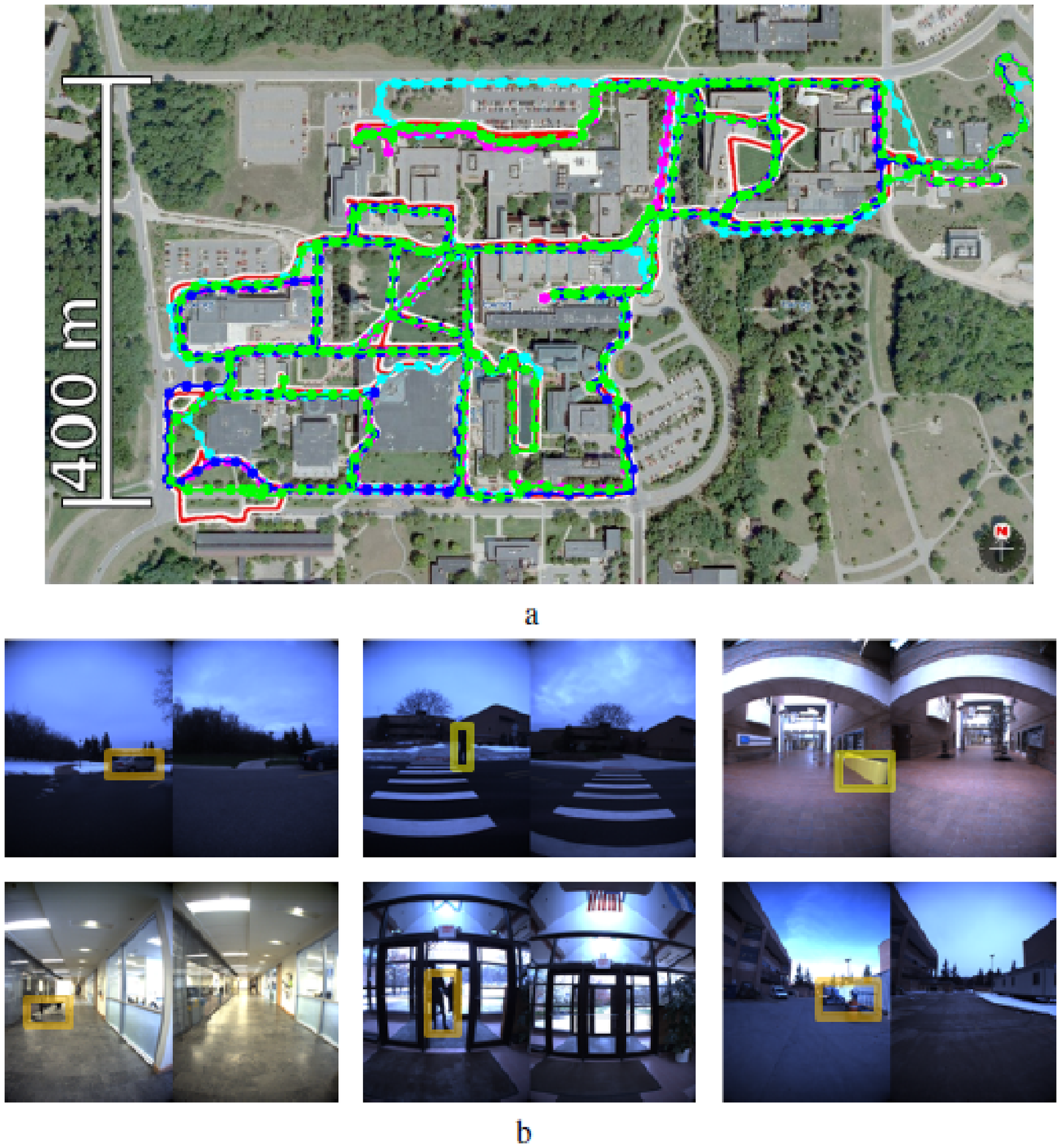}{}
\caption{
Scalable change detection for long-term map maintenance. (a) Experimental environment. The trajectories of the four datasets ``2012/01/22," ``2012/03/31," ``2012/08/04," and ``2012/11/17" used in our experiments are visualized in green, purple, blue, and light-blue curves and overlaid on the bird's-eye-view imagery obtained from NCLT dataset \cite{1}. (b) Examples of changes. For each panel, the left and right figures are a query image and its corresponding reference (i.e., mapped) image, respectively.}\label{fig:C}
\end{center}
\vspace*{-8mm}
\end{figure}
}

\newcommand{\figD}{
\begin{figure}[t]
  \begin{center}
\FIG{8.5}{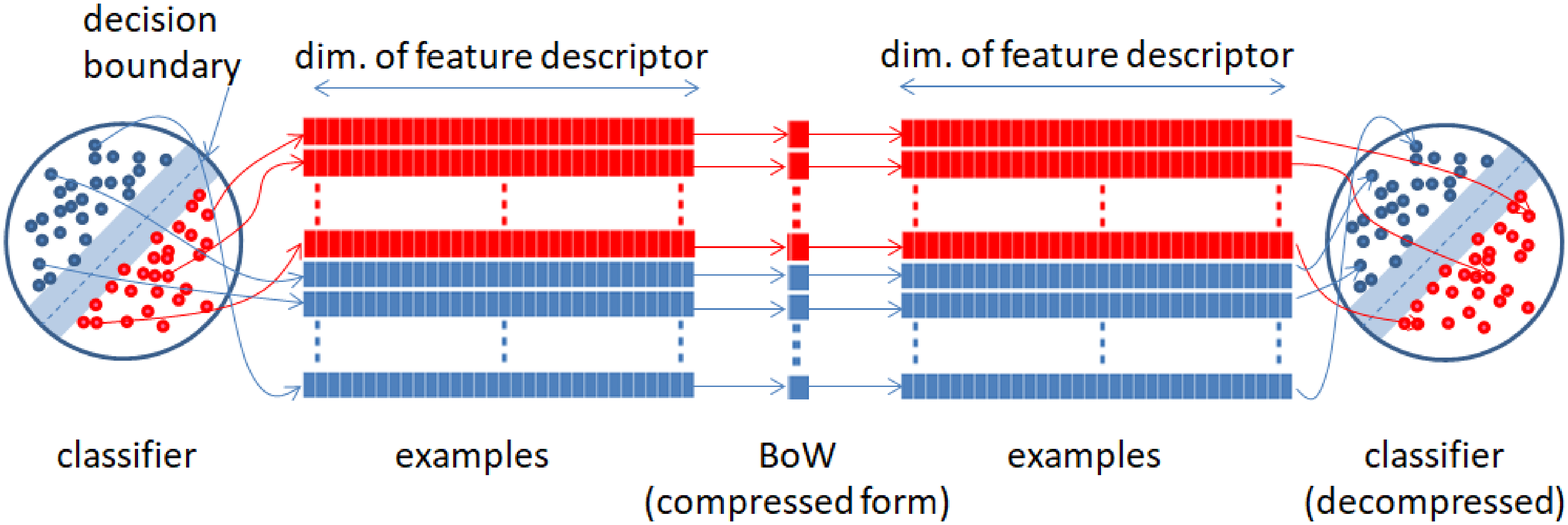}{}
\caption{Pipeline of classifier-compression/-decompression task. Red: positive examples. Blue: negative examples.}\label{fig:D}
\vspace*{-7mm}
  \end{center}
\end{figure}
}

\newcommand{\figF}{
\begin{figure}[t]
  \begin{center}
\FIGR{8}{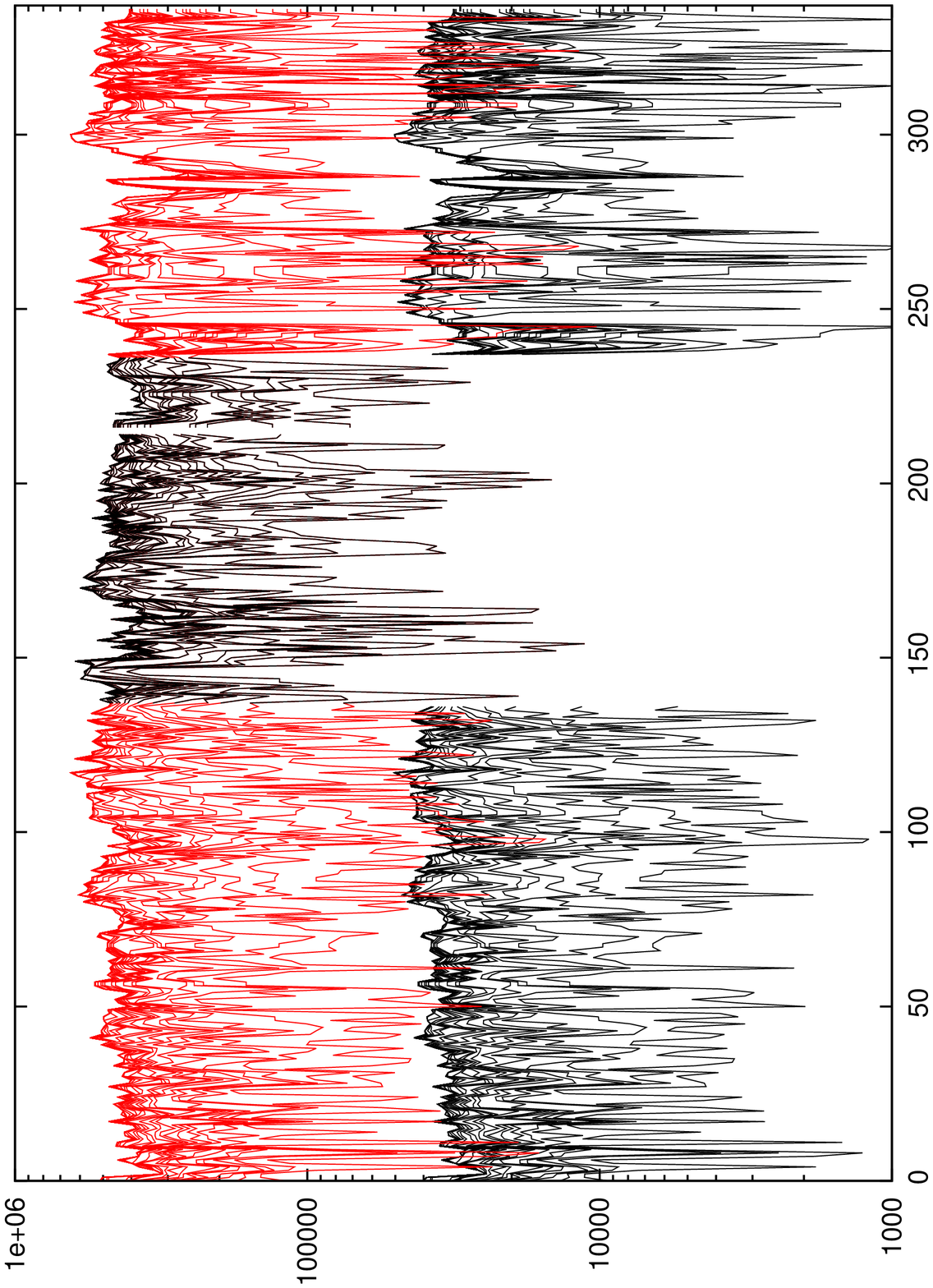}{}\vspace*{-3mm}\\
\caption{Effect of map compression. Vertical axis: Total space cost (bits) for all the place-specific change classifiers. Horizontal axis: Frame ID. Black and red curves exhibit space costs for compressed and non-compressed (or decompressed) maps. Each $k$-th curve corresponds to the $k$-th object cluster, and the intervals between the $(k-1)$-th and $k$-th curves equals the space cost of the $k$-th object cluster.}\label{fig:F}
\vspace*{-3mm}
  \end{center}
\end{figure}
}

\newcommand{\figG}{
\begin{figure*}[t]
  \begin{center}
\FIG{16}{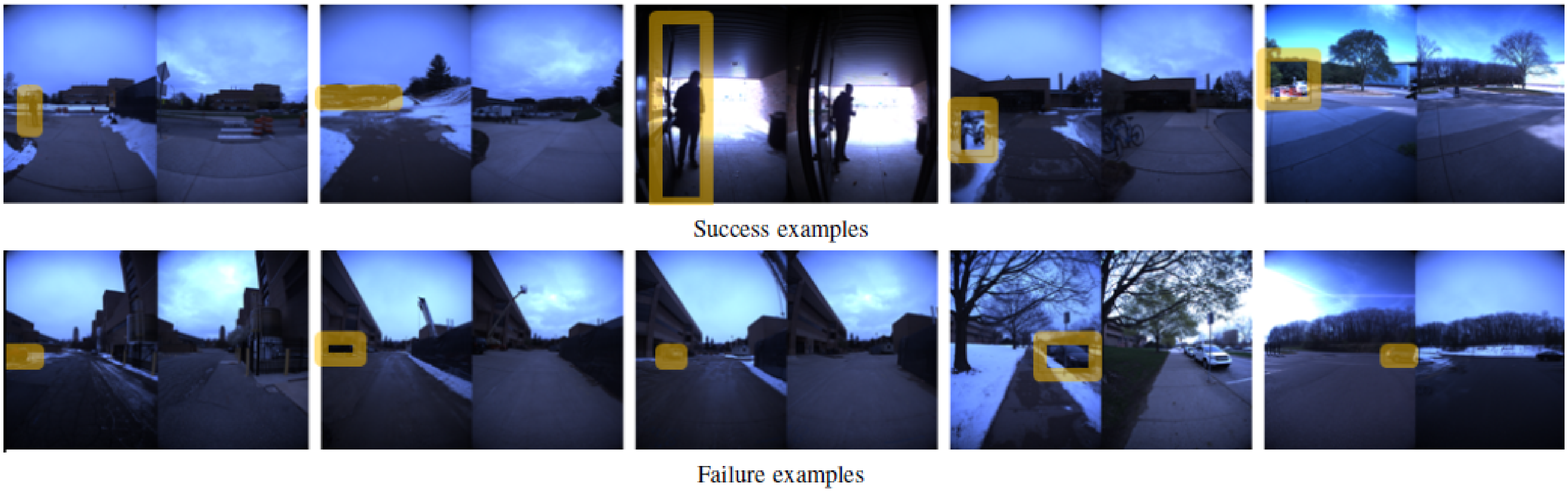}{}
\caption{Change detection examples.}\label{fig:G}
  \end{center}
\end{figure*}
}

\newcommand{\figH}{
\begin{figure}[t]
  \begin{center}
\FIG{8.5}{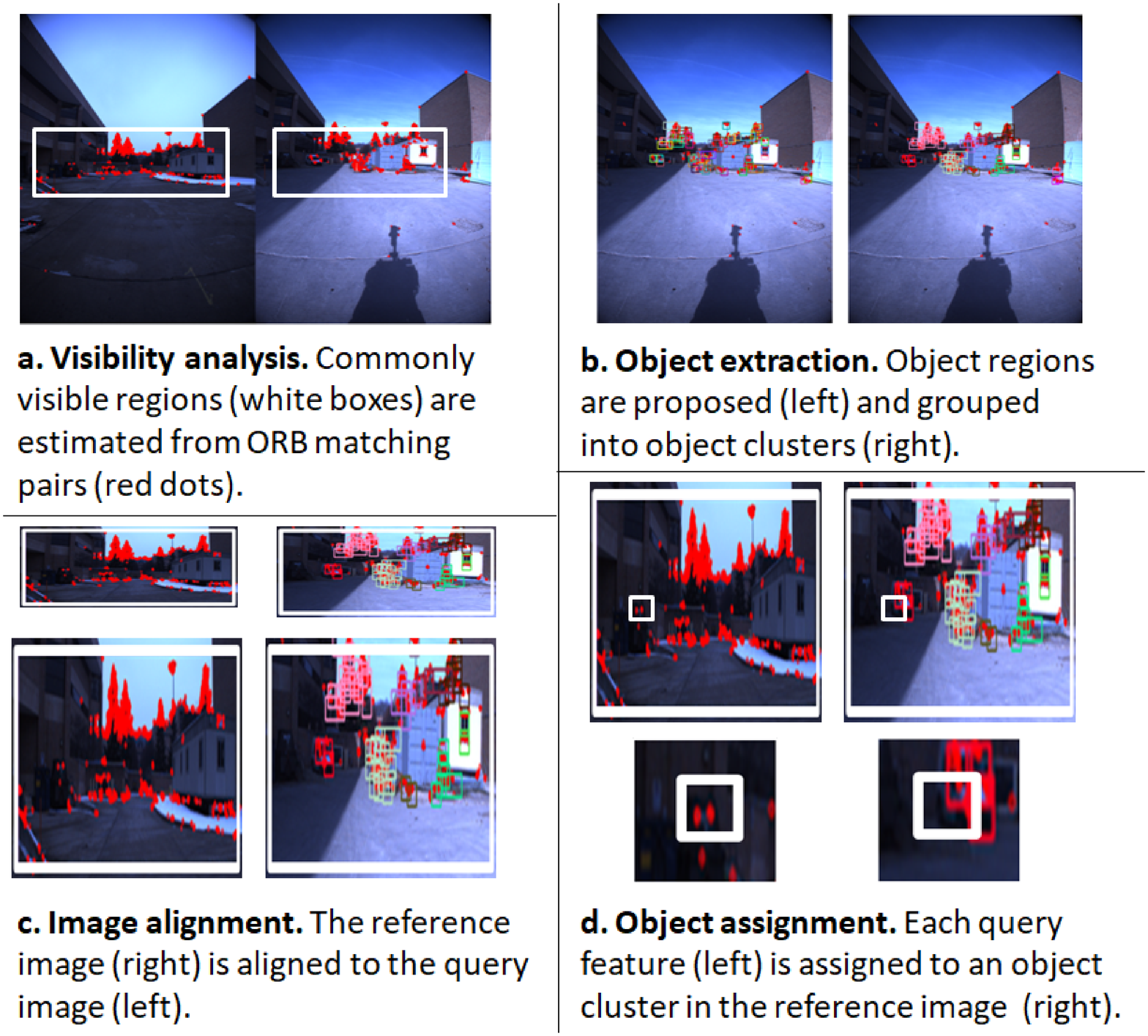}{}\vspace*{-2mm}\\
\caption{Overview of scene recognition.}\label{fig:H}
\vspace*{-5mm}
  \end{center}
\end{figure}
}

\title{\LARGE \bf
Zero-Shot Learning to Manage a Large Number of Place-Specific Compressive Change Classifiers
}

\author{Tanaka Kanji
\thanks{Our work has been supported in part by 
JSPS KAKENHI 
Grant-in-Aid for Young Scientists (B) 23700229,
for Scientific Research (C) 26330297,
and for Scientific Research (C) 17K00361.}
\thanks{The authors are with Graduate School of Engineering, University of Fukui, Japan. 
{\tt\small tnkknj@u-fukui.ac.jp}}}

\maketitle

\begin{abstract}
\SW{
With recent progress in large-scale map maintenance and long-term map learning, the task of change detection on a large-scale map from a visual image captured by a mobile robot has become a problem of increasing criticality. Previous approaches for change detection are typically based on image differencing and require the memorization of a prohibitively large number of mapped images in the above context. In contrast, this study follows the recent, efficient paradigm of change-classifier-learning and specifically employs a collection of place-specific change classifiers. Our change-classifier-learning algorithm is based on zero-shot learning (ZSL) and represents a place-specific change classifier by its training examples mined from an external knowledge base (EKB). The proposed algorithm exhibits several advantages. First, we are required to memorize only training examples (rather than the classifier itself), which can be further compressed in the form of bag-of-words (BoW). Secondly, we can incorporate the most recent map into the classifiers by straightforwardly adding or deleting a few training examples that correspond to these classifiers. Thirdly, we can share the BoW vocabulary with other related task scenarios (e.g., BoW-based self-localization), wherein the vocabulary is generally designed as a rich, continuously growing, and domain-adaptive knowledge base. In our contribution, the proposed algorithm is applied and evaluated on a practical long-term cross-season change detection system that consists of a large number of place-specific object-level change classifiers.
}{

}
\end{abstract}

\section{Introduction}

\noeditage{
\figC
}

\SW{
This paper considers the problem of scalable change detection  using  a  novel, compact  representation  of an  environment map (Fig. \ref{fig:C}). With recent progress in large-scale map maintenance \cite{2} and long-term map learning \cite{3}, the task of change detection on a large-scale map, from a visual image captured by a mobile robot, has become a problem of increasing criticality \cite{4}. We  aim to enhance the scalability from the novel perspective of map compactness while maintaining the effectiveness of the change detection system.
}{
}

\SW{
In general, change detection has the goal of detecting changes between a robot's view image and a previously constructed map. A major challenge in change detection is to effectively address appearance variations of the change objects. As the variations are inherently place-specific \cite{5} and attributed to various factors (e.g., objects, viewpoint, background, illumination conditions, occlusions), it is challenging to obtain a general-purpose change model. Till the present, the most fundamental scheme reported for addressing this challenge is to directly compare each view image against the corresponding reference (i.e., mapped) image using handcrafted features \cite{6} or deep-learning techniques \cite{4d}. However, this fundamental scheme requires explicit memorization of every possible mapped image and exhaustive many-to-many image comparisons, which severely limits the scalability in time and space.
}{
}

\SW{
Our approach is inspired by the recent success in the community of change-classifier-learning \cite{7,8,8d,9}. Instead of memorizing the large collection of mapped images, the change-classifier-learning approach simply learns an essential change classifier that is often significantly compact and nonetheless exhibits generalization capability. In \cite{7}, Gueguen et al. presented a change detection method for overhead imagery wherein they train a support vector machine (SVM) with linear kernel with L1-regularization and L2-loss as a semi-supervised scene-specific change classifier from manually labeled positive examples (i.e., changes) and plenty of available negative examples (i.e., no-changes). Moreover, it is preferable to use only a single change classifier; however, it is challenging for a single classifier to capture the place-specific variations of changes \cite{5} or to flexibly incorporate the latest local changes in the map \cite{10}. Therefore, we opt to use multiple place-specific change classifiers, each of which is responsible for each specific place region and the place-specific change classifiers can be flexibly and locally updated, added, or deleted at a marginal fixed cost. Each classifier receives local feature-descriptors (e.g., ORB descriptors) as input and evaluates their likelihood of change. However, it is not straightforward to apply the typical framework of change-classifier-learning to our application domain of autonomous robotics, wherein we are not provided examples with labels (i.e., change/no-change). Instead, the mapper-robot itself should collect examples in an unsupervised manner.
}{
}

\SW{
In this study, we address the above issue from the viewpoint of zero-shot learning (ZSL) \cite{11d}. ZSL is a domain-adaptation technique in the field of machine learning and was originally proposed as a strategy to obtain classifiers for arbitrary, novel categories when no labeled example is available. In particular, we are inspired by mining-based ZSL \cite{11}, wherein training examples for novel unseen categories (in the target domain) are collected by mining an external knowledge base (EKB) such as a search engine. To realize our ZSL-based change-classification framework, it is necessary to address the following questions: (1) How does one train classifiers? (2) How does one represent examples? (3) How does one prepare an EKB? 
}{
}

\SW{
The key concepts required for answering these questions are as follows:
\begin{enumerate}
\item We train classifiers with the latest examples when necessary to incorporate the latest changes in the map;
\item We use bag-of-words (BoW) as a compact and discriminative representation of examples;
\item We employ a visual word vocabulary as the EKB,
\end{enumerate}
which are inspired by the use of BoW as a compressed form of place-specific classifier in our previous study \cite{23}.
}{
}
\SW{
The above concepts exhibit several advantages: (1) We only have to memorize training examples rather than the classifier themselves because typical classifier-learning algorithms such as SVM \cite{12} and nearest neighbor (NN) \cite{13} allow us to reproduce
a classifier, given the same training examples \cite{23}. (2) We can incorporate the latest changes into the classifiers by simply adding or deleting a few training examples that correspond to these classifiers, while the training-time overhead per place can be reasonably low by using efficient training algorithms such as NN and SVM. (3) We can represent each training example in a significantly compact form of a visual word \cite{23}, implying that the training example is approximated by its nearest neighbor exemplar feature from a visual word vocabulary. (4) We can share the vocabulary with other related task-scenarios including BoW-based self-localization, SLAM, lifelong learning, and open-set recognition, wherein the vocabulary is generally designed as a rich, continuously growing, and domain-adaptive knowledge base. In experiments, the proposed framework is evaluated in a cross-season change detection setting using the publicly available NCLT dataset \cite{1}.
}{
}

\subsection{Relation to Other Works}

\SW{
A majority of the present state-of-the-art change detection systems are based on image differencing or pairwise image comparison \cite{5}. 
In \cite{26},
a scene alignment method for image differencing
is proposed based on ground surface reconstruction, texture projection, image rendering, and registration refinement.
In \cite{4}, a deep deconvolutional network for pixel-wise change detection was trained and used for comparing query and reference image patches.

Some recent studies use change-classifier-learning to realize more efficient and compact change detection \cite{7,8,8d,9}. Our algorithm is inspired by these classifier-learning approaches and advances a step further from the perspective of unsupervised ZSL. The work in \cite{7} can be considered as one of the most relevant works to our study. In their work, change detection in overhead imagery is addressed. A BoW model with tree-of-shape features is employed to achieve more effective accuracy-efficiency tradeoff. Based on these features, linear canonical correlation analysis is employed to learn a subspace to encode the notion of change between images. To reduce the cost of label acquisition by human photo-interpreters, a semi-supervised SVM framework is introduced. In contrast, we address unsupervised settings, wherein examples have to be labeled automatically by the robot itself rather than by human labelers.
}{
}

\SW{
Our study focuses on a monocular camera as the sole input device. This is a significantly challenging setting compared with other change detection settings, which assume a more richer information source, including 3D model \cite{14}, stereo or image sequence \cite{6}, 3D data \cite{15d}, multi-spectral overhead imagery \cite{16}, and object model \cite{ren2017faster}.
}{
}

\SW{
In the area of field robotics, there is substantial work on change detection systems for patrolling \cite{17}, agriculture \cite{18}, tunnel inspection \cite{18d}, and damage detection \cite{7}. We consider that our extension would also contribute to these applications from the novel perspective of map compactness.
}{
}

\SW{
This work is a part of our study on compact map model for scalable change detection and long-term map maintenance \cite{acpr17kanji,23,21,icra04kanji,icra15kanji}. In previous studies, two scenarios have been considered. One is the ``global localization" scenario \cite{icra04kanji}, wherein the change detection system is required to work under global viewpoint uncertainty. As this scenario requires access to the entire large-size maps in memory, individual images were used in their compressed form of bag-of-words \cite{acpr17kanji}. In contrast, the current study focuses on the alternative ``pose-tracking" scenario, wherein the system can assume that the robot's viewpoint is tracked over time. A similar setting is addressed in our recent study in \cite{21} with a key variation being that classifiers are not compressed. As the pose-tracking scenario requires access to only a marginal portion of the submaps that correspond to the robot's surroundings, we are allowed to use mapped images in a less-compact form of the decompressed classifiers. Nevertheless, it is necessary to maintain the other classifiers compressed. Hence, the efficiency of compression/decompression is another critical topic, as also discussed in this paper.
}{

}

\section{Approach}

\SW{
Our goal is to enhance the compactness of map representation for scalable change detection. As mentioned, this rules out typical image-differencing approaches that require memorization of a number of high-resolution mapped images proportional to the map size. Therefore, we decide to use the change-classifier approach. More specifically, we employ multiple place-specific change classifiers to capture the place-specific nature of changes and to enable flexible and local update of the change model. The key concept is to represent each classifier by their training examples and compress each training example to a visual word; this enables a significantly compact and domain-adaptive change model.
}{

}

\SW{
Based on the above consideration, we use a set of place-specific change classifiers and represent each by a BoW. Following literature \cite{bow}, a BoW is represented as an unordered collection of visual words, or visual features vector quantized by a codebook of exemplar features called vocabulary.
In our approach, an $i$-th classifier is represented by positive (i.e., change) and negative (i.e., no-change) sets of visual words, $S^+_i$ and $S^-_i$, respectively. Each visual word $w\in S^{\{+,-\}}_i$ is represented in the form:
\begin{equation}
w=
\langle
w^a, w^{r}
\rangle.
\end{equation}
$w^a$ is a $B$-bit code called appearance word and is an identifier for its nearest neighbor local-feature-descriptor in the vocabulary. We employ ORB feature descriptor \cite{orb} as local-feature-descriptor because it is a rapid binary keypoint descriptor that has been applied to numerous real-time computer-vision and robotic-mapping problems \cite{orbslam}. The number of ORB descriptors per image is set to 2000. We denote the appearance-word vocabulary as $V$ (i.e., $B=\log_2 |V|$). $w^{r}$ is called pose word and represents the spatial location of the feature keypoint descriptor with respect to the object region; it incurs a small constant space cost $B'$=$\log_2|A|$ (bits) where $|A|$ is the area [pixels] of the object region. Consequently, the total space-cost (bits) of our BoW-based model approximately sums up to
\begin{equation}
C = \sum_{x\in X} \left[ \sum_{r\in R_x} \left[ \sum_{w\in W_r} \log_2|V| + B' \right] \right],
\end{equation}
where $X$ is the set of place regions, $R_x$ is the object regions
(described in \ref{sec:learning})
in the $x$-th place, and $W_r$ are the visual words belonging to the object $r$.
}{
}

\SW{
Apparently, it is necessary to minimize the sizes of $X$, $R_x$, $W_r$ and $V$ without compromising change-classification accuracy. Joint minimization of these is generally intractable, and in this study, we straightforwardly consider each minimization problem separately. Minimizing $R_x$, $W_r$ and $V$ are the tasks of change-aware object proposal, feature extraction, and vocabulary design. We have also addressed the issue of vocabulary design and the use of BoW as a compressed form of classifiers in an alternative context of visual robot localization in \cite{23}. Minimizing $P$ is called unsupervised place-definition and workspace-partitioning discovery. We have addressed this problem in our previous studies in the context of place recognition \cite{mva17kanji} and change recognition \cite{21}. In this study, we straightforwardly partition the entire sequence of mapped images in the workspace into equal-length subsequences, each of which corresponds to each place class.
}{
}

\noeditage{
\figD
}

\noeditage{
\figF
}

\SW{
Our algorithm consists of two distinct phases: training and classification. The training procedure runs through the following steps: (t1) Train classifiers. (t2) Represent classifiers by their positive/negative examples. (t3) Approximate the examples by BoWs. (t4) Memorize the examples in the form of BoW. The classification procedure runs through the following steps: (c1) Lookup the vocabulary to determine examples that correspond to the specified reference image. (c2) Reproduce classifiers from the examples. (c3) Classify query features using the reproduced classifiers. (c4) Delete classifiers. Note that steps (c2) and (c4) function as decompression and compression of classifiers, respectively (Fig. \ref{fig:D}). The time overhead of the compression task is more or less negligible, while that of the decompression task is marginal albeit noticeable. Therefore, it is necessary to perform task scheduling of decompression task adequately. In this study, we follow a straightforward strategy: When the robot approaches or enters a new submap's region, decompress the classifiers of its surrounding submaps. 

Figure \ref{fig:F} demonstrates our strategy. In the figure, the frame IDs ranging from 137 to 237 correspond to the classifiers for the robot's surroundings. As illustrated, the space cost of the compressed classifiers is significantly lower than that of non-compressed or decompressed classifiers, which demonstrates the effectiveness of our approach.
}{
}

\subsection{Classifier-learning}\label{sec:learning}

\SW{
The place-specific change model aims to detect changes at the level of object. It consists of a set of object-level classifiers, each of which aims to learn the appearance of a known object (i.e., no-change object) in the reference image of a specific place and then, to classify an unseen query feature as either change or no-change with respect to the learned object. To this end, object segmentation both at training and at classification tasks significantly influences the performance of the object-level change detection.
}{
}

\noeditage{
\figH
}

\SW{
The training phase begins with the extraction of a collection of object proposals from the specified reference image (Fig. \ref{fig:H}b left). We use BING object-proposal algorithm \cite{bing} to achieve this because it is highly efficient to provide object proposals with category-independent image windows. Then, the proposed object-regions are further grouped into clusters of spatially-near regions, termed object clusters (Fig. \ref{fig:H}b right). We use a criterion in this clustering wherein any two object regions belong to the same cluster if their bounding boxes overlap each other. 
Then, an additional ``background" object cluster, 
which treats the entire image region as the object region, is added.
Then, a change classifier is trained for each object cluster. We will implement and evaluate two types of classifiers: nearest neighbor, or SVM, and four types of SVM kernel: linear, sigmoid, polynomial, or RBF. The above methods of object proposal and classifier-learning are detailed below.
}{
}

\SW{
The BING algorithm generally produces a large number of object proposals (e.g., 2$\times 10^3$ proposals per image). The proposals generally contain numerous false positives. Evaluating all the proposals is computationally intractable. Therefore, we decide to select a marginal portion of the object proposals. From the viewpoint of effectiveness of our BoW model, we evaluate spatial density of visual words $N/A$, where $A$ and $N$ are the area of the object proposal's bounding box and the number of visual words inside the bounding box, respectively. Then, we select 400 object proposals with highest density. Subsequently, near-duplicate object regions that overlap significantly with object regions with higher density are eliminated. We consider two object regions $i$ and $j$ to be in the near-duplicate condition if their overlap ratio 
$A_{ij}/\min(A_i, A_j)$
exceeds a threshold of 0.5; here, $A_i$, $A_j$ and $A_{ij}$ are the areas [pixels] of the objects $i$ and $j$ and their overlap region.
}{
}

\SW{
The change classifier addresses one-sided classification problem, wherein only negative examples (i.e., no-change) are provided during the training stage. In order to obtain a sufficiently large number of positive examples, we consider the task that we call change mining \cite{21}, the aim of which is to mine features of potential change objects from a large feature collection of the EKB. In this study, we consider three strategies for change mining: ``uniform," ``farthest," and ``nearest." The ``uniform" strategy uniformly samples positive examples from the EKB, while the ``farthest" and ``nearest" strategies sample positive examples that are respectively the farthest and nearest in Hamming distance from the specified negative examples. For all the strategies, each example in the EKB that is nearer than 10 bit from its nearest neighbor negative is considered inappropriate as a positive example and eliminated from the candidates of positive examples prior to the change mining. 
To enable efficient training, the number of training examples 
is truncated to a maximum number 400. 
}{
}

\subsection{Change Classification}\label{sec:classification}

\SW{
The change classification process aims to rank local feature-descriptors in query images in the order of likelihood of changes. It consists of two distinct steps: (1) image registration and (2) change ranking. Both the steps are detailed below.
}{

}

\SW{
The image registration step, which aims to align query and reference images into the same coordinate frame (Fig. \ref{fig:H}c), is a necessary pre-processing step for a substantial majority of change detection tasks \cite{5}. Note that image alignment from monocular image-pairs is significantly ill-posed, particularly when we are provided only BoW representations of the images. We have tested three strategies for feature matching---ORB keypoint matching with and without post-verifications using RANSAC \cite{24} and using vector field consensus (VFC) \cite{25}---and observed that these two post-verification strategies (i.e., RANSAC and VFC) are not adequately stable in our scenario of highly-complex scenes. Therefore, we use ORB keypoint matching as a method for image alignment.
}{
}

\SW{
In this study, we follow a fundamental procedure for image registration \cite{26}, which assumes a linear transformation from reference- to query-image coordinate (Fig. \ref{fig:H}c),
although a key difference is that we are given the BoW representation instead of raw feature descriptors. Note that the transformation algorithm requires pairs of feature keypoints matched between query and reference images. In order to filter out outlier matches to the maximum, an image region commonly visible in both images is estimated in the form of bounding box (Fig. \ref{fig:H}a). More formally, keypoints of matched visual words $M$ are sorted in ascending order of $x$- or $y$-coordinates, and we define $\lfloor \delta|M| \rfloor$-th and $\lceil (1-\delta)|M| \rceil$-th elements ($\delta=0.1$) in the sorted lists as the $x$- and $y$-locations of the boundary of the visible regions.
}{

}

\SW{
Two types of additional spatial cues are obtained as a byproduct of the image registration. The first one is global spatial-information for visibility analysis (Fig. \ref{fig:H}a), wherein local features outside the commonly visible region are not considered as change-object candidates. The second is local spatial-information by which each object region (i.e., bounding box) in the reference image is first transformed to the query image's coordinate (Fig. \ref{fig:H}c); following this, the query image's local features outside the transformed bounding box (Fig. \ref{fig:H}d) are not considered as matching candidates of the specific object in the reference image. Taking into account the transformation errors, we expand the transformed bounding box by $\Delta L$ [pixel], wherein the margin $\Delta L$ is currently set to 10\% of the image width.
}{

}

\SW{
The change-ranking step aims to rank query features by the likelihood of change, given the learned place-specific object-level change classifiers. First, each query feature is assigned to the spatially-near object cluster in the corresponding reference image (Fig. \ref{fig:H}d). We assign each query feature to an object cluster if its keypoint is located within one of the object cluster's bounding box. Then, each query feature is input to the assigned object cluster's change classifier, and the classifier outputs the probability $p_r$ of the query feature not originating from the learned object cluster $r$. We employ Platt scaling \cite{27} with five-fold cross-validation to convert an output value of the SVM classifier to the probability estimate or use a Gaussian $[1-\exp(-d^2/\sigma_d^2)]$ to convert an output distance of the nearest neighbor classifier $d$ into a probability value. Given the output probability value $p_c$ from each object cluster $c$, the probability of the query feature of interest being changed is computed by
\begin{equation}
p=\min_{c} p_c.
\end{equation}
}{

}

\noeditage{
\figB
}

\SW{
In order to render the set of top-ranked features more diverse, we introduce the idea of non-maximal suppression and penalize features that are considered similar to a higher-ranked feature. In our view, two features are similar if they belong to the same object cluster. Based on the concept, we extract object clusters from a specified query image and group query features into clusters in a similar manner as in the reference images; then, features in each cluster are ranked in descending order of the probability $p$. Then, the entire set of query features is re-ranked in the ascending order of augmented score $r+(1-p)$, where $r(\in[1, C])$ is the abovementioned intra-cluster rank, $C$ is the size of the largest cluster, and $p(\in [0,1])$ is the original probability-estimate.
}{

}

\SW{
Note that all query features do not always belong to object clusters, and there may be isolated query features that do not belong to any object cluster. We empirically determined that such isolated features are generally less-reliable. Therefore, we assign them a worst intra-cluster rank $r=C+1$. 
}{

}

\section{Experiments}

\SW{
We evaluated various change detection strategies using NCLT dataset \cite{1}. NCLT dataset is a long-term autonomy dataset for robotics research collected on the University of Michigan's North Campus. The dataset consists of omnidirectional imagery, 3D lidar, planar lidar, GPS, and odometry, and we use the monocular images from the front-directed camera (``camera \#5") for our change detection tasks. Fig. \ref{fig:B} shows some examples of scene recognition.

During the vehicle's travel through both indoor and outdoor environments (Fig. \ref{fig:C}a), it encounters various types of changes, which originate from the movement of individuals, parking of cars, furniture, building construction, opening/closing of doors, and placement/removal of posters (Fig. \ref{fig:C}b). There are also nuisance changes that originate from illumination alterations, viewpoint-dependent changes of objects' appearances and occlusions, weather variations, and falling leaves and snow. A critical and significant  challenge in a substantial majority of change detection tasks is to discriminate changes of interest from nuisances. This renders our change detection task significantly more demanding.
}{

}

\noeditage{
\figA
}

\SW{
We use four datasets ``2012/1/22," ``2012/3/31," ``2012/8/4," and ``2012/11/7" that correspond to four sessions of vehicle navigation. These datasets consist of 5095, 3994, 4877, and 5118 images of size 1232 $\times$ 1616. 
We manually created 7571 pairs of corresponding query and reference images, which correspond to ``revisiting" or ``loop-closing" situations \cite{icra04kanji}, using the global viewpoint information. While 548 of the pairs are change image-pairs, 6714 pairs are no-change image-pairs; the remaining 309 pairs are 
not independent of the 548 change pairs and are not used in the experiments. We categorized small changes (e.g., 10$ \times $10 [pixels]) that typically originated from distant objects, into no-change because it is challenging to detect such small changes by visual change detection algorithm. As a result, such small objects are likely to cause pseudo false-positive detection by change detection algorithms. We use specific combinations of query and reference image sets: (query, reference) = (2012/1/22, 2012/3/31), (2012/3/31, 2012/8/4), (2012/8/4, 2012/11/7), or (2012/11/7, 2012/1/22). The vocabulary is sized 1 M (i.e., 20-bit visual words) by default, and its exemplar features are randomly sampled from the reference set's visual features. Change objects in the 548 change image pairs are manually annotated in the form of bounding boxes. 
}{
}

\SW{
The average time-overhead for compression and decompression per classifier was 0.3 s and time-overhead per place was 3.0 s using a non-optimized implementation of SVM from SVM light \cite{joachims1999svmlight} on a laptop PC (Intel Core i5-4200 2.5 GHz). This implies that the compression/decompression can function in real-time when the robot moves at the speed of 1 m/s and the workspace is divided into 3-m-length places. The total travel distance was 22,181 m. 
}{
}

\SW{
We consider a straightforward change detection task scenario: Given a collection of 100 query images acquired by a robot, identify the image that is the changed one and the locations of the change features within that image. The size of an image collection is set to 100, and it consists of a change image and 99 random no-change images. We created 548 collections based on the above 548 change image pairs, for each of which 99 no-change images are randomly sampled from the 6714 no-change pairs, and the collections are commonly used as dataset by each change detection algorithm.
}{
}

\SW{
We consider various strategies of change mining, classifiers, and image registration. For the change mining, we implemented three methods: ``uniform," ``farthest," and ``nearest," which are described in II-A. For the classifier, we implemented the SVM classifier with four kernels---linear, sigmoid, polynomial, and RBF---and the nearest neighbor classifier with Hamming distance metric. For the image registration, we aim to verify the efficacy of the visibility analysis strategy presented in II-B and implemented the common visible region in the form of bounding box (``horizontal-vertical") as well as two additional methods for visibility analysis. One of these two is a method that omits $y$-direction boundaries, which is motivated by an observation that $y$-direction spatial information is generally unreliable (``horizontal"). The other is a method that completely omits the bounding-box information, which corresponds to not using the visibility analysis strategy (``none").
By default, we use SVM with RBF kernel as the classifier, uniform sampling with 10-bit distance threshold as the change mining strategy, and the 20-bit vocabulary.
}{
}

\SW{
Figures \ref{fig:A}, \ref{fig:Aa}, \ref{fig:Ab}, \ref{fig:Ac} and \ref{fig:Ad} exhibit the performance results. We evaluated the various methods over the independent 548 image collections. Each image collection consists of 100 pairing of query and reference images, and each query image contains 2000 features to score. We assess whether the change detection task on an image collection is successful or not and compute the success ratio over all the image collections. For the assessment, all the $100$$\times$$2,000$$=$$200,000$ features are sorted in the ascending order of the augmented score (described in \ref{sec:classification}), and if a ground-truth change feature is ranked within the top-$X$ (\%) with respect to the sorted list, the change detection is considered as success; otherwise, it is considered as failure. We performed the evaluation for various $X$ values: 
0.1\%, 0.25\%, 0.5\%, 1\%, 2.5\%, and 5\%, 
which respectively correspond to the top 
200, 500, 1000, 2000, 5000, and 10000
ranked query features.
}{
}

\SW{
Figure \ref{fig:A} illustrates results of evaluating the fundamental effects of the proposed strategies---object-level change detection (\ref{sec:learning}) and non-maximal suppression (\ref{sec:classification}). We developed a comparison method termed ``non-object," which employs a single image-level classifier rather than the object-level classifier. The image-level classifier is implemented as a single object-level classifier that treats the entire image as the object region. We also developed another comparison method termed ``non-suppression," which does not employ the non-maximal suppression technique. In the figure, ``object" indicates the proposed method (classifier: SVM w/ RBF kernel, vocabulary size 20 bit), and ``non-object" and ``non-suppression" indicate the two comparison methods. It can be observed that the proposed ``object" method evidently outperforms the other two, and it is verified the object-level change classifiers are substantially more powerful than the comparison feature-level methods.
}{
}

\SW{
Figure \ref{fig:Aa} illustrates the result of comparing classifiers and kernels. We evaluated different classifiers and kernels (nearest neighbor, SVM with linear, sigmoid, polynomial, and RBF kernels). It can be seen that SVM with RBF kernel outperforms the other classifiers and kernels.
}{
}

\noeditage{
\figAa
\figAb
\figAc
\figAd
\figG
}

\SW{
Figure \ref{fig:Ab} shows a result of comparing different change mining strategies. 
We evaluated different change mining strategies (uniform with 10-bit, 20-bit, and 30-bit distance threshold, farthest, and nearest). It can be seen that uniform sampling with 10-bit distance threshold outperforms the other methods. The reason might be that the uniform sampling strategy tends to produce a more diverse set of positive examples than other strategies, and thus the resulted classifier exhibits adequate generalization capability.
}{
}

\SW{
Figure \ref{fig:Ac} illustrates the result of evaluating the visibility analysis strategies. We evaluated different visibility analysis strategies (horizontal-vertical, horizontal, and none). 
It can be observed that the horizontal-vertical strategy outperforms the other two strategies when the parameter $X$ is sufficiently marginal.
}{
}

\SW{
Figure \ref{fig:G} illustrates examples of successful and failed change detection. Here, we used the abovementioned default settings. As illustrated, changes originating from moving objects such as cars and pedestrians are generally easy to detect owing to the fact that their visual appearances are significantly discriminative from other background objects such as roads and trees. However, the detection of moving objects is rendered a substantially challenging task when a similar moving object appears in the corresponding reference image also (Fig. \ref{fig:G} failure examples). Although spatial information (e.g., location and size) of such similar objects are generally dissimilar to that of the query object, they are ordinarily accepted as matching candidates of the query object because the search region is expanded to overcome coordinate transformation error (as explained in \ref{sec:classification}). Other challenging cases include change objects such as boxes and tables (Fig. \ref{fig:G} failure examples) whose visual appearance are significantly similar to background objects such as floors and walls.
}{
}

\section{Conclusions}

\SW{
We presented a change detection framework that realizes map compactness while maintaining detection efficiency. Rather than memorizing pre-trained classifiers, our ZSL-based approach has to memorize only the compact indexes to their training examples that are mined from EKB. Experimental results on place-specific object-level change classifiers have demonstrated high potential. The proposed algorithm is efficient and very simple to implement. It should therefore be convenient to integrate it into existing frameworks of change detection (e.g., change detection in 3D, stereo images) to enhance their compactness.
}{
}

\bibliographystyle{IEEEtran}
\bibliography{cd}

\end{document}